\documentclass[conference]{IEEEtran}
\ifCLASSINFOpdf \else \fi
\usepackage{color}
\usepackage{graphicx}
\usepackage{subfigure}
\usepackage{hyperref}
\hypersetup{colorlinks=true,linkcolor=red,anchorcolor=red,citecolor=green}
\usepackage{amsfonts}
\usepackage{mathrsfs}
\usepackage{makecell}

\usepackage[cmex10]{amsmath}
\usepackage{amssymb}
\usepackage[]{hyperref}
\usepackage{booktabs}

\usepackage{colortbl}
\definecolor{mygray}{gray}{.92156}

\usepackage{multirow}
\usepackage{CJK}
\usepackage{algorithm}
\usepackage{algorithmic}   

\usepackage{array}
\usepackage{url}
 \usepackage[numbers,sort&compress]{natbib}

\begin{document}
\title{LOGO: Video Text Spotting with Language Collaboration and Glyph Perception Model}


\DeclareRobustCommand*{\IEEEauthorrefmark}[1]{%
    \raisebox{0pt}[0pt][0pt]{\textsuperscript{\footnotesize\ensuremath{#1}}}}
\author{
	\IEEEauthorblockN{
		Hongen Liu\IEEEauthorrefmark{1}, 	
		Di Sun\IEEEauthorrefmark{2}, 
		Jiahao Wang\IEEEauthorrefmark{1},
            Yi Liu\IEEEauthorrefmark{3}, 
		Gang Pan\IEEEauthorrefmark{1, *}} 
\IEEEauthorblockA{\IEEEauthorrefmark{1}College of Intelligence and Computing, Tianjin University}
\IEEEauthorblockA{\IEEEauthorrefmark{2}Tianjin University of Science and Technology} 	\IEEEauthorblockA{\IEEEauthorrefmark{3}Baidu Inc.}

 \IEEEauthorblockA{\{hongenliu, wjhwtt, pangang\}@tju.edu.cn, dsun@tust.edu.cn, liuyi22@baidu.com, }
}

\maketitle

\begin{abstract}
\par 
\textnormal{
Video text spotting (VTS) aims to simultaneously localize, recognize and track text instances in videos. 
To address the limited recognition capability of end-to-end methods, recent methods track the zero-shot results of state-of-the-art image text spotters directly, and achieve impressive performance. However, owing to the domain gap between different datasets, these methods usually obtain limited 
 tracking trajectories on extreme dataset. Fine-tuning transformer-based text spotters on specific datasets could yield performance enhancements, albeit at the expense of considerable training resources. In this paper, we propose a \textbf{L}anguage C\textbf{o}llaboration and \textbf{G}lyph Percepti\textbf{o}n Model, termed \textbf{LOGO}, an innovative framework designed to enhance the performance of conventional text spotters. To achieve this goal, we design a language synergy classifier (LSC) to explicitly discern text instances from background noise in the recognition stage.  Specially, the language synergy classifier can output text content or background code  based on the legibility of text regions, thus computing language scores. Subsequently, fusion scores are computed by taking the average of detection scores and language scores, and are utilized to re-score the detection results before tracking. By the re-scoring mechanism, the proposed LSC facilitates the detection of low-resolution text instances while filtering out text-like regions. Moreover, the glyph supervision is introduced to enhance the recognition accuracy of noisy text regions. In addition, we propose the visual position mixture module, which can merge the position information and visual features efficiently, and acquire more discriminative tracking features. Extensive experiments on public benchmarks validate the effectiveness of the proposed method.}
\end{abstract} 



\IEEEpeerreviewmaketitle
\section{Introduction}
\par Video text spotting (VTS) involves the concurrent localization, tracking, and recognition of text instances within videos. This task has constituted a fundamental task in various domains, such as driver assistance \cite{Roadtext1k}, video understanding \cite{video_understand}, and video retrieval \cite{video_retrieval}. Compared to text spotting in static images, video text spotting presents heightened challenges due to external interference (e.g., motion blur, out-of-focus, occlusions, text-like regions, variations in illumination, and perspective shifts).
\par Recent methods \cite{transdetr,cotext} aim to unify the training of all modules within an end-to-end framework, and to incorporate temporal information across multiple frames. These methods have yielded significant improvements over traditional tracking-by-detection pipelines \cite{edit_distance,transpotter,iou_based_track,Zuo_etal}. However, despite these advancements, the architectures of these methods are not optimally tailored for text instances, and the spotting models lack pre-training on extensive text datasets, such as Synth150K \cite{ABCNet} and Synth800K \cite{SythText}. Consequently, the performance of text spotting on single frame remains constrained, particularly in complex scenarios. To address the limited recognition capability of end-to-end methods, GoMatching \cite{GoMatching} designs the Long-Short Temporal Matching-based tracker (LST-Matcher) to track the zero-shot results of the state-of-the-art (SOTA) image text spotter DeepSolo \cite{deepsolo}. Notably, GoMatching surpasses end-to-end methods \cite{transdetr,cotext} by a considerable margin in terms of both spotting performance and training cost.

\begin{figure}
\centering
\subfigure[]{
\includegraphics[width=0.24\textwidth]{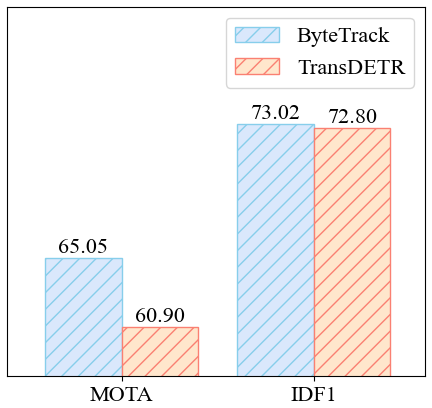}}
 \hspace{-0.15in}
\subfigure[]{
\includegraphics[width=0.24\textwidth]{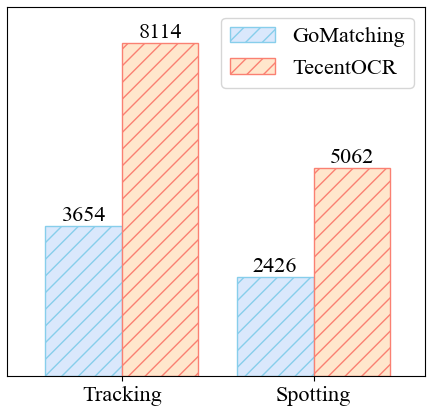}}
\caption{The rethink to advantages and limitations of GoMatching \cite{GoMatching}. 
(a) The video text spotting performance of ByteTrack \cite{bytetrack} and TransDETR \cite{transdetr} on ICDAR2015 Video. The spotting results of single frame in ByteTrack \cite{bytetrack} are acquired using DeepSolo \cite{deepsolo}, while MOTA \cite{MOTA} and IDF1 \cite{IDF1} serve as the chosen evaluation metrics.
(b) The number of most tracked trajectories from GoMatching \cite{GoMatching} and TencentOCR for video text tracking and video text spotting on DSText.}
\label{fig:subfig} 
\end{figure}

\par Since GoMatching \cite{GoMatching} achieves impressive performance, we revisit its design paradigm, and analyze its advantages and limitations. As depicted in Fig. \ref{fig:subfig} (a), even the conventional tracker ByteTrack \cite{bytetrack} instead of LST-Matcher is employed to associate text instances from DeepSolo \cite{deepsolo}, the MOTA and IDF1 metrics outperform the best end-to-end method TransDETR \cite{transdetr}. This result underscores the significance of robust image text spotter. However, relying solely on the zero-shot results of DeepSolo may constrain tracking trajectories, particularly in more intricate video text datasets. As depicted in Fig. \ref{fig:subfig} (b), due to the high proportion of small and dense text instances on DSText \cite{DSText}, the tracked trajectories of GoMatching encompass less than half of those achieved of the leading method TencentOCR. Undoubtedly, the limited  tracked trajectories result in the potential semantic information loss, and hinder the better understanding for videos. To address this challenge, fine-tuning DeepSolo on specific datasets could yield performance enhancements, albeit at the expense of considerable training resources. Considering the pivotal role of synergy between the detection and recognition modules in transformer-based text spotters, an alternative approach involves enhancing the spotting performance of conventional detectors through the integration of a synergy module.
\par
\begin{figure}
\centering
\subfigure[PP-YOLOE-R]{
\begin{minipage}[b]{0.47\textwidth}
\includegraphics[width=1\textwidth]{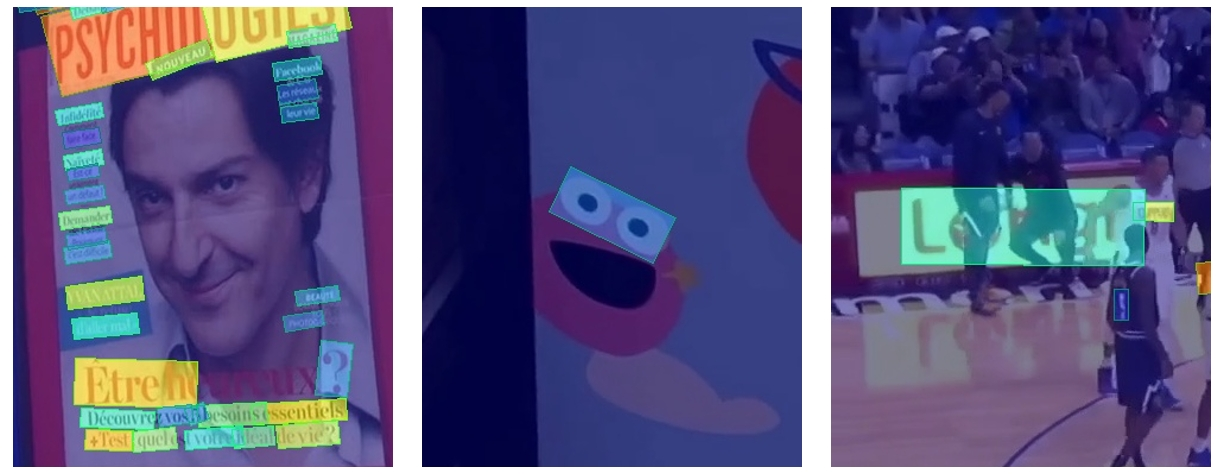}
\end{minipage}
}
\subfigure[Language synergy classifier]{
\begin{minipage}[b]{0.47\textwidth}
\includegraphics[width=1\textwidth]{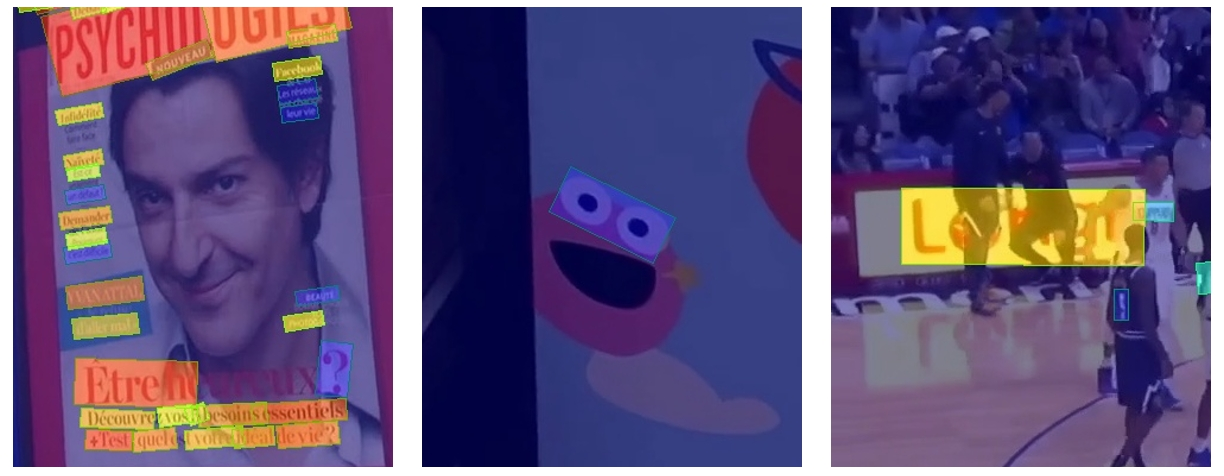}
\end{minipage}
}
\caption{ Visualization of the detection results from PP-YOLOE-R and the re-scored results from Language Synergy Classifier. (a) PP-YOLOE-R. (b) Language Synergy Classifier (LSC). Confidence scores range from 0 to 1, with colors closer to red indicating higher confidence scores.} \label{intro2}
\end{figure}

 To this end, an efficient YOLO based model PP-YOLOE-R \cite{ppyolor} is selected as text detector to improve training efficiency. The detection results are visualized in Fig. \ref{intro2}.
We argue that the poor performance of traditional detectors stems from the ambiguity inherent in visual features. Primarily, the high resolution representation of small texts often occur in low-level feature map, and the semantic information deficiency within low-level feature maps often leads to their misclassification as background. Furthermore, visual features may inadvertently incorporate background noise when text instances are subjected to physical phenomena such as occlusion, motion blur, or perspective variation. As illustrated in the first and third columns of Fig. \ref{intro2} (a), partial text instances with low resolution or affected by various physical phenomena may receive low scores, and be entirely overlooked. Secondly, visual features solely capture contextual and structural information of text, and lack linguistic knowledge to discern the legibility of text instances. Even though the text-like regions contravene language rules, the network inevitably detects them. As shown in the second column of Fig. \ref{intro2} (a), the eyes of a red cartoon face is erroneously detected as a text instance due to their structural resemblance to the word `OO'. 

\par In this paper,  we propose a \textbf{L}anguage C\textbf{o}llaboration and \textbf{G}lyph Percepti\textbf{o}n Model, termed \textbf{LOGO}. Initially, a language synergy classifier (LSC) is designed to explicitly differentiate text instances from background noise in the recognition stage. Since YOLO based detectors typically generate numerous proposals, training the detector and synergy module simultaneously produces expensive computational overhead. Therefore, the LSC is trained offline using the detection outcomes of PP-YOLOE-R. Technically, ground truths are selected as positive samples, while prediction boxes with low intersection over union with ground truths serve as negative samples. Next, the text regions of these samples are extracted from the original image using RoI Rotate operations, and the recognition results of positive samples and negative samples are respectively encoded as text codes and background codes. During inference, the LSC outputs text content
or background code based on the legibility of text regions, thus computing language scores.  After that, the fusion scores are computed by taking the average of language scores and detection scores, and are utilized to re-score the detection results prior to tracking. As depicted in Fig. \ref{intro2} (b), this approach enables the detection of text instances with low resolution while disregarding text-like regions. Furthermore, the glyph supervision is introduced to enhance the network's perception to glyph structures of text instances, thereby improving recognition accuracy in noisy text regions. Additionally, to make up the absence of query features \cite{deepsolo} in PP-YOLOE-R, the Visual Position Mixture Module (VPMM) is proposed. This module can merge position information from the detector and visual features from the LSC to obtain more discriminative tracking features. Finally, the LST-Matcher in GoMatching is employed to associate text instances across multiple frames. The main contributions are summarized as follows:
\begin{itemize}
    \item We offer a new perspective into video text spotting. The synergy between visual information and linguistic knowledge in LOGO alleviates the ambiguity of visual features, allowing the model to distinguish text instances from background noise more accurately.  
    \item  We introduce the glyph supervision to improve the recognition accuracy of noisy text regions, and devise the visual position mixture module to acquire more discriminative tracking features. 
    \item Extensive experiments on public benchmarks for three video tasks demonstrate the effectiveness of the proposed method.
\end{itemize}

\section{Related Work}

\subsection{Text Reading in Single Image}
\par Early methods \cite{textboxes, east, RRPN, RRD, RethinkMask,PixelLink, PSENet, DBNet, FCENet,CRNN, CTC2, CTC3,TextScanner,show_atten,AON,ESIR,recognition_work1,ABINet,glyph_rec} to text reading are primarily designed for text detection and text recognition, separately. In text detection, traditional methods \cite{textboxes, east, RRPN, RRD, RethinkMask} typically follow object detection pipelines, and regard text detection as a bounding box regression problem. However, these regression-based methods aim to enhance the detection performance of rotated texts, and often struggle with arbitrarily-shaped texts. To address this limitation, segmentation-based methods \cite{PixelLink, PSENet, DBNet} and  control points methods \cite{ABCNet, FCENet} are proposed. For text recognition, current methods can be divided into two categories: language-free and language-aware methods. Language-free methods \cite{CRNN, CTC2, CTC3,TextScanner, glyph_rec} view text recognition as character classification task, and recognize text by visual feature. In contrast, language-aware methods \cite{show_atten,AON,ESIR,recognition_work1, ABINet} typically utilize attention models to implicitly learn the relationship between characters, and achieve promising results especially in scenarios with complex visual cues.

\par Recently, there has been a trend among researchers to integrate text detection and recognition into end-to-end trainable networks \cite{FOTS,ABCNet,ROISpotter1,textdragon}. These approaches leverage RoI-Align and its variants in the detection module to extract text features, and employ a recognition module to learn linguistic information from visual features. Benefiting from the unified architecture, these methods can  significantly enhance inference speed and overall performance. 
\par While the RoI-based text spotters successfully combine the two tasks into a unified model, the intrinsic synergy between detection and recognition has not been sufficiently explored. To address this, several methods \cite{textspottertransform,SWinTextSpotter,deepsolo,estextpotter} propose various synergy strategies to eliminate extra connectors, such as RoI-Align. For instance, CharNet \cite{charnet} and MaskTextSpotter \cite{masktextspotter} replace the recognition module by a character segmentation branch, and adopt different channels to represent corresponding English characters. TESTR \cite{textspottertransform} integrates the two tasks with dual decoders: a location decoder for detection and a character decoder for recognition. SwinTextSpotter \cite{SWinTextSpotter} introduces recognition conversion to guide text localization in the detection module, and suppress noise background in the recognition feature. Based on a DETR-like baseline, DeepSolo \cite{deepsolo} detects and recognizes text simultaneously using a single decoder with explicit points solo. ESTextSpotter \cite{estextpotter} employs a task-aware decoder with a vision-language communication module to achieve the interaction between two tasks. However, the synergy mechanisms in these methods remain unexplained, and the recognition module cannot explicitly participate in the detection process. To address limitations, we propose a novel language synergy classifier. This classifier explicitly distinguishes between background noise and text contents, and re-scores the detection results based on the legibility of text regions.


\subsection{Text Reading in Videos}
\par Compared to text reading in single image, research on text reading in videos is relatively scarce. These researches primarily focus on video text tracking and video text spotting.
\par For video text tracking, traditional methods \cite{mineto,edit_distance,iou_based_track, Zuo_etal} typically adhere to the tracking-by-detection pipeline, and employ hand-crafted features (e.g., HOG, IoU, edit distance, etc.) to associate text instances across consecutive frames. However, hand-crafted features often exhibit weaknesses in complex scenarios, and result in limited tracking performance. To address this challenge, Feng et al. \cite{Semantic} proposes an appearance-semantic-geometry descriptor to track text instances with appearance changes. Additionally, TextSCM \cite{TextSCM} utilizes a siamese complementary module and a text similarity learning network to respectively relocate missing text instances and distinguish text with similar appearances.
\par For video text spotting, earlier methods \cite{edit_distance,transpotter,iou_based_track} often require intricate pipelines, and involve separate training for detection, tracking, and recognition modules. To alleviate computational costs and expedite inference speed, YORO \cite{YORO} and its enhanced version Free \cite{Free} integrate text recommenders to select high-quality texts from streams, thereby recognizing selected texts only once instead of frame-by-frame. CoText \cite{cotext} introduces a lightweight video text spotter, and leverages contrastive learning to model long-range dependencies across multiple frames. Following the paradigm of MOTR \cite{MOTR}, TransDETR \cite{transdetr} optimizes all modules within a unified framework. However, the multi-task optimization in TransDETR may lead to conflicts. Furthermore, due to lack of well-designed structures for text instances in its architecture, TransDETR exhibits limited recognition capabilities for single frame, compared to advanced image text spotters \cite{deepsolo}. To address these limitations, GoMatching \cite{GoMatching} introduces a re-scoring mechanism and a long-short term temporal matching module to transform off-the-shelf query-based image text spotters into video text spotters. By this design, it saves on training budget and achieves notable performance. Nonetheless, relying solely on the zero-shot results of state-of-the-art image text spotters may result in limited trajectories, particularly in challenging datasets \cite{DSText}. Moreover, since state-of-the-art image text spotters typically adopt transformer-based architectures, training them on specific datasets can be costly. In contrast, we select the YOLO based detector to improve training efficiency, and enhances its performance by introducing a synergy module. Besides, the visual position mixture module is proposed to acquire more discriminative tracking features. These features can replace the query features in DeepSolo \cite{deepsolo}, and facilitate a more effective video text spotting process.
\section{Methodology}

\begin{figure*}
  \centering
  \includegraphics[width=7in]{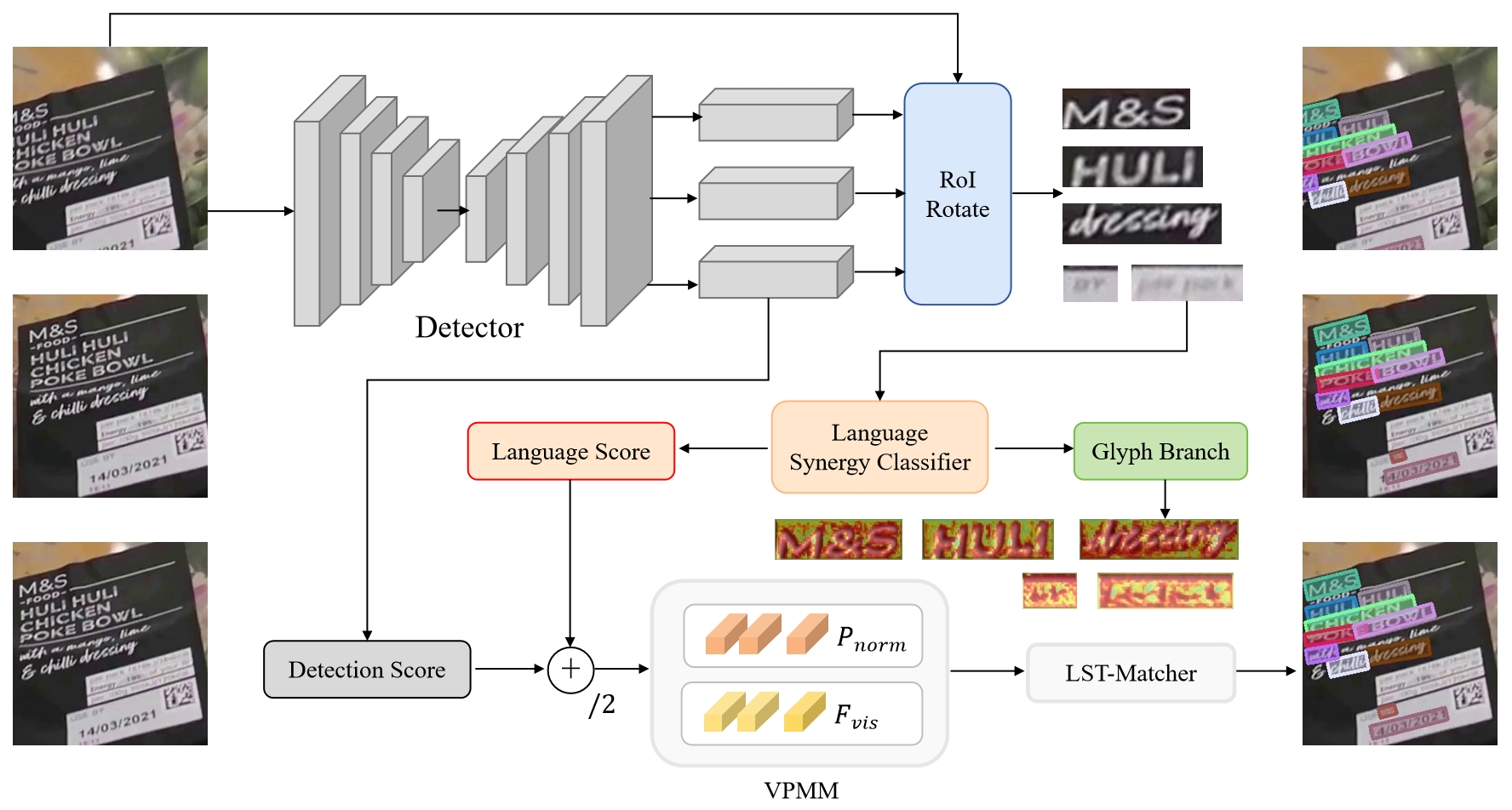}
  \caption{The overall pipeline of LOGO. In this pipeline, the YOLO based rotated detector is employed to improve training efficiency, and the Language synergy classifier (LSC) is designed to distinguish text instances from background noise based on language knowledge. Simultaneously, we
  introduce the glyph branch to enhance the model's perception for text structure, and propose the Visual Position Mixture Module (VPMM) to fuse the position information and visual features of the detection results. The fusion features will be fed into LST-Matcher for video text tracking. }\label{overview}
\end{figure*}

\subsection{Overview}
\par The overall pipeline of LOGO is depicted in Fig. \ref{overview}. Initially, the input image is processed by the detector to yield the detection results. After that, detection outcomes exceeding the predefined score threshold are selected and their corresponding text regions in the original image are extracted via RoI Rotate operation. These text regions are then fed into the Language Synergy Classifier (LSC) for the calculation of language score. Subsequently, the language scores and the detection scores are combined using weighted fusion to derive fusion scores. The prediction results, now corrected by the fusion scores, are input into  Visual Position Mixture Module (VPMM) to integrate position information and visual features, and generate tracking features. These tracking features are then sent to the LST-Matcher to produce the VTS results. Additionally, the Glyph Supervision is designed to enhance the model's perception of text structure, thereby improving text recognition accuracy in noisy environments.

\subsection{Language Synergy Classifier}
 \begin{figure*}
  \centering
  \includegraphics[width=7in]{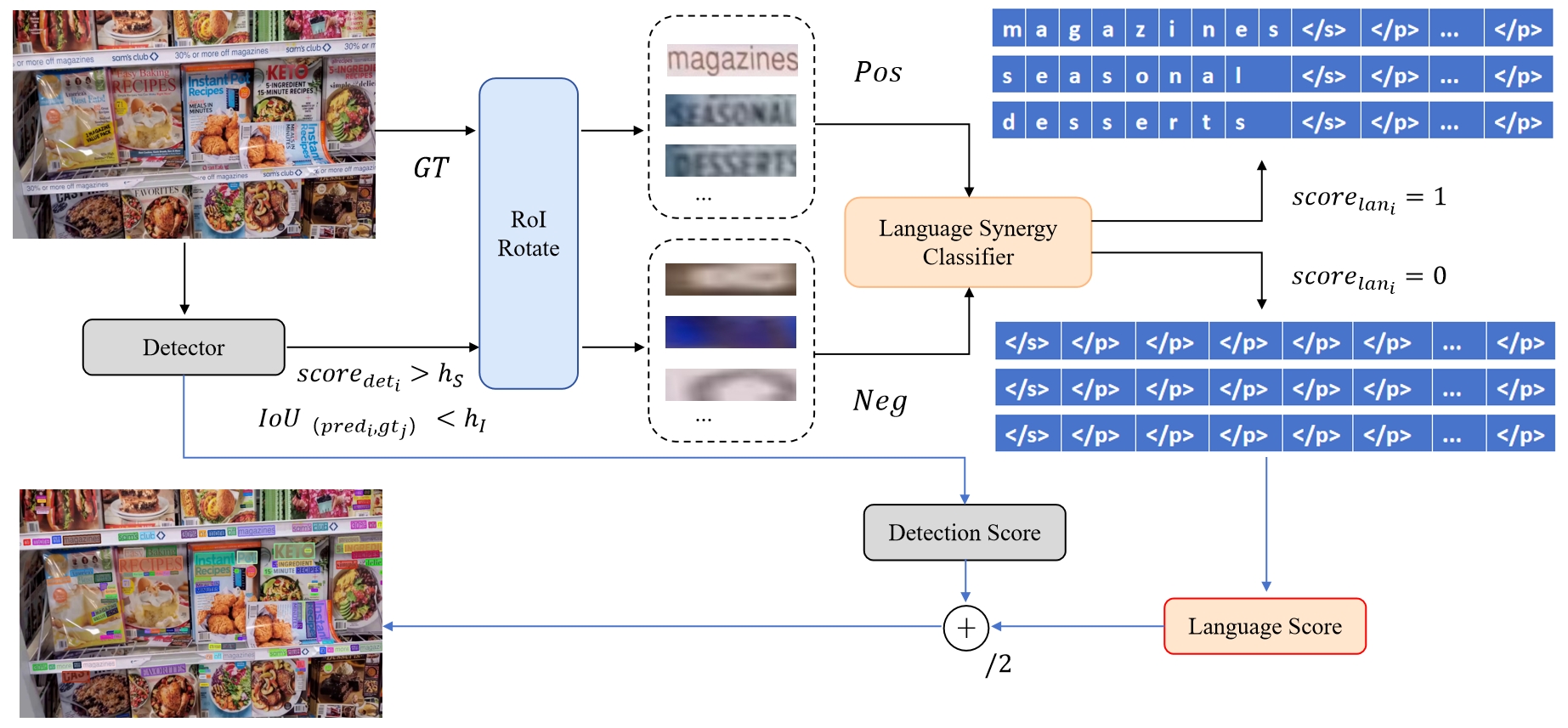}
  \caption{The architecture of language synergy classifier. In the training phase, ground truths (GT) and prediction boxes with low Intersection over Union (IoU) with GT are defined as positive and negative samples, respectively. Next, we extract the text regions of these samples by RoI Rotate operation, and encode their recognition results. Then, these text regions and recognition encodings are fed into LSC for network optimization. In the inference phase, the language synergy classifier outputs language scores based on the recognition results. After that, the fusion scores are computed by taking the average of language scores and detection scores, and are utilized to recalibrate the detection results. Notably, ``GT'' represents the ground truths from training datasets, while $\langle /s \rangle$ and $\langle /p \rangle$ represent the end symbol and padding symbol, respectively. }\label{LSC}
\end{figure*}
\par To facilitate explicit participation of the recognition module in the text detection process, we propose the Language Synergy Classifier (LSC). When the detection results are legible, the LSC module generates specific text content. Conversely, if the detection results indicate background noise, the module produces background code.
\par The architecture of the Language Synergy Classifier (LSC) is delineated in Fig. \ref{LSC}. Given that anchor-free detectors typically yield numerous proposals, training the synergy module online akin to transformer-based text spotters presents challenges. Therefore, the detection results from the rotated detector are leveraged to train the LSC offline. Subsequently, ground truths (GT) and prediction boxes with low Intersection over Union (IoU) with GT are respectively chosen as positive and negative samples based on Eq. \ref{neg_sample}. 
\begin{multline}\label{neg_sample}
N e g =\left\{{pred_{i}} \mid \operatorname{IoU}_{\left({pred }_i, gt_j\right)}<h_I\right.,\\
score\left._{{det }_i}>h_S, i \leq m, j \leq n, i \in N^{+}, j \in N^{+}\right\}
\end{multline}
where the $Neg$ represents the negative sample set. $pred_{i}$ and $gt_{j}$ denote the $i$-th detection result and the $j$-th ground truth, respectively. $h_I$ and $h_S$ represent the IoU threshold value and score threshold value used for selecting detection results. $score_{det_{i}}$ denotes the confidence score of the $i$-th detection result. $m$ and $n$ denote the number of detection results and ground truths, respectively.
\par In scenarios with a high proportion of small and dense text, directly sampling RoI features from the detector may introduce background noise, and lead to potential performance degradation. Consequently, the text regions are extracted from the original images using the RoI Rotate operation to preserve high resolution features. Subsequently, these sampled regions are inputted into the Language Synergy Classifier (LSC) module to learn the language representation, and the corresponding recognition results are encoded. Specially, positive samples are encoded as text codes, terminated by the end symbol $\langle EOS \rangle$, and padded with the symbol $\langle PAD \rangle$. Similarly, Negative samples are encoded with the end symbol $\langle EOS \rangle$, and padded with the symbol  $\langle PAD \rangle$. The detailed encoding process can be defined as follows:
\begin{align}
&Encode_{{pos}_{ik}}=\begin{cases}{\operatorname{Code}\left[s_{k}\right],} &   0 \leq k \leq { len }_s -1 \\ {EOS}, &  k = { len }_s \\ 
{ PAD}, & { len }_s<k \leq { len }_{ {max }}-1\end{cases} \\
&Encode_{{neg}_{jk}}=\begin{cases}{EOS} ,&   k=0  \\ { PAD},  & 1 \leq k \leq { len }_{ {max }}-1\end{cases}
\end{align}
where $Encode_{{pos}_{ik}}$ and $Encode_{{neg}_{jk}}$ represent the $k$-th character code for $i$-th positive sample and $j$-th negative sample, respectively. $EOS$ and $PAD$ denote the end symbol and padding symbol, respectively. $s_{k}$ refers to the $k$-th character of the current word, while ${ len }_s$  and ${ len }_{max}$ denote the length of the current word and the maximum encoding length, respectively. The ``Code" variable represents the encoding dictionary for digits and English characters.
\par In the inference phase, the LSC module computes language scores based on the recognition results. Specifically, these language scores are assigned a value of 1 if the recognition results are text contents, and 0 if they are identified as background noises. This assignment can be formally expressed as:
\begin{align}
   { score }_{ {lan }_i}=\left\{\begin{array}{l}
1, \quad  { code }_{{lan }_{i}}[0] \neq  { EOS } \\
0, \quad \operatorname{code}_{ {lan }_{i}}[0]= { EOS }
\end{array}\right.
\end{align}
where ${ score }_{ {lan }_i}$ represents the language score for the $i$-th samples, and ${ code }_{{lan }}[0]$ denotes the $0$-th code of the $i$-th text instances.
\par Ultimately, the fusion scores are computed by taking average of the detection scores and language scores, and are utilized to recalibrate the detection results prior to tracking. Consequently, the scores of text-like regions are attenuated, while the scores of legible text at low resolutions are amplified. 
\subsection{Glyph Supervision}
\par 
Due to the intricate visual interference in complex scenarios, video text spotters often struggle to accurately recognize text regions amidst noisy backgrounds. Motivated by \cite{glyph_rec}, we introduce the glyph supervision to aid the model in learning glyph representations of text foregrounds, thus getting rid of interference from background noise.
\par
\begin{figure}
  \centering
  \includegraphics[width=3.2in]{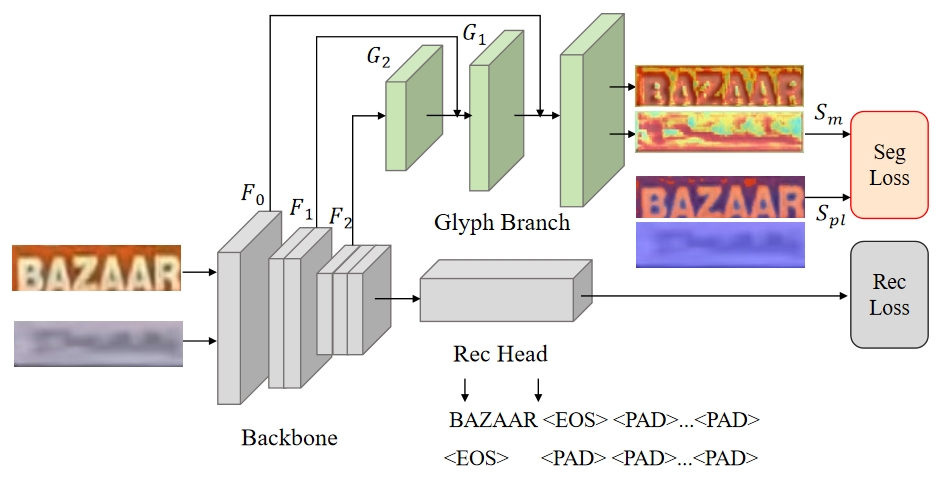}
  \caption{The structure of glyph supervision. $F_{0}$, $F_{1}$, $F_{2}$ represent the features extracted from the backbone, while $G_{0}$ and $G_{1}$ denote the features from the glyph branch. Additionally, $S_{m}$ and $S_{pl}$ signify the segmentation masks and pseudo-labels, respectively.}
  \label{glyph supervision}
\end{figure}

 As depicted in Fig. \ref{glyph supervision}, the K-means algorithm is employed to generate pseudo-labels $S_{pl}$ for text masks, and the glyph branch is integrated into the language synergy classifier. This branch is utilized to learn representations of text foregrounds. Specifically, backbone features $F_{0}$, $F_{1}$, and $F_{2}$ are leveraged to generate glyph features $G_{1}$, and $G_{2}$, and the final text segmentation mask $S_m$ is derived through Eq. \ref{glyph_seg}.
\begin{align}\label{glyph_seg}
\left\{\begin{array}{l}
G_2=\varphi\left(F_2\right), \\
G_1=\varphi\left(\mathcal{U}\left(G_2, s_1\right)+F_1\right), \\
S_m=\tau\left(\mathcal{U}\left(G_1, s_0\right)+F_0\right),
\end{array}\right.
\end{align}
where $\varphi(\cdot)$ denotes one convolutional layer followed by BatchNorm and ReLU activation functions. $\tau(\cdot)$ denotes three convolutional layers followed by ReLU activation functions. Additionally, $\mathcal{U}(\cdot)$ represents the 2$\times$ upsampling operation for $G_{k}$ with a resolution of $s_{k}$ (i.e., $H_k$ $\times$ $W_k$). 
\par
\begin{figure}
  \centering
  \includegraphics[width=3.2in]{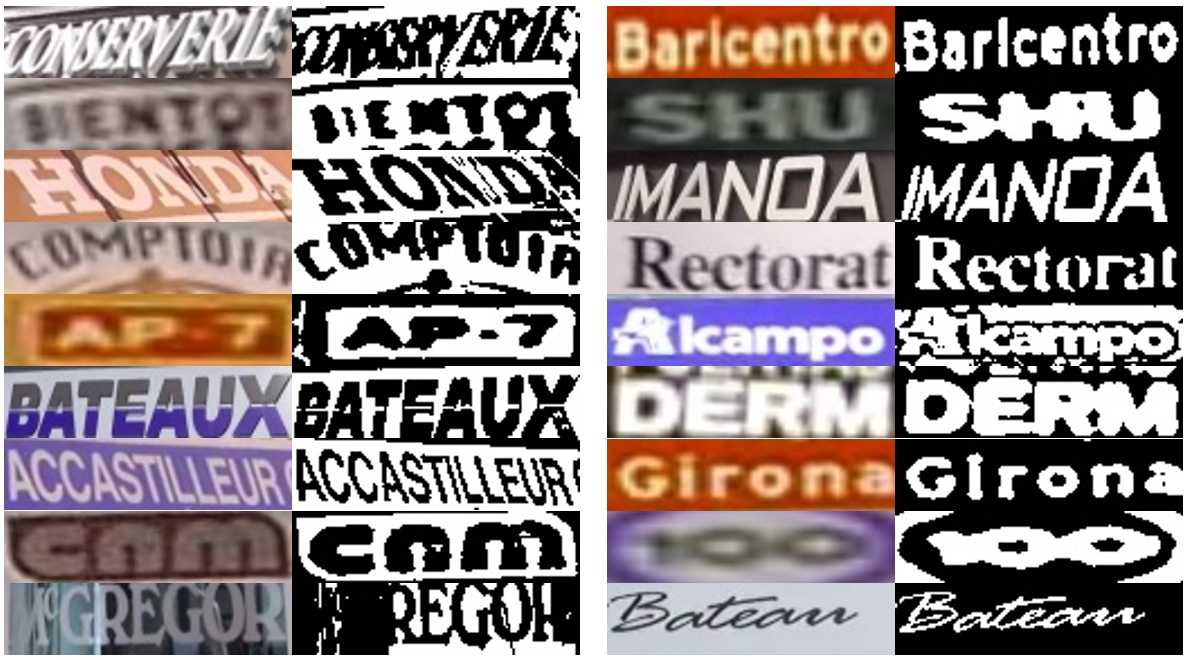}
  \caption{Pseudo-labels generated by K-means}
  \label{pesudo_label}
\end{figure}

 However, due to the constraint of limited prior knowledge, the pseudo-labels $S_{pl}$ generated by the K-means algorithm exhibit instability. As depicted in Fig. \ref{pesudo_label}, the labels of text foregrounds fluctuate between 0 and 1. Because of the ambiguity of labels assignment, the network's ability to distinguish text foreground from background noise is greatly hindered. To address this issue, the Mean Square Error (MSE) loss is adopted as the optimization criterion for the glyph branch instead of the Binary Cross Entropy (BCE) loss employed in \cite{glyph_rec}. The MSE loss function can be formally defined as follows:
\begin{align}\label{glyph_segloss}
\mathcal{L}_{s e g}=\left\|S_m-S_{p l}\right\|_2^2
\end{align}

\subsection{Visual Position Mixture Module}
\par
\begin{figure}
  \centering
  \includegraphics[width=3.2in]{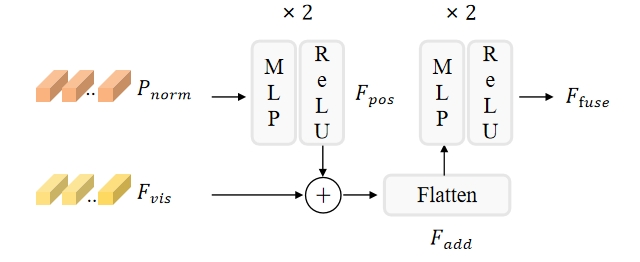}
  \caption{The structure of visual position mixture module (VPMM). $F_{norm}$ and $F_{vis}$ represent the normalized centerline vertexes and visual features, respectively. The combination of  Multi-Layer Perceptron (MLP) and ReLU activation functions is utilized for feature projection, while  $\times 2$ signifies the number of the combination layers. 
  }
  \label{VLMM}
\end{figure}

 To compensate the absence of query feature in YOLO based detector, we propose the Visual Position Mixture Module (VPMM). This module can integrate position information from the detector and visual features from the Language Synergy Classifier (LSC) module, thereby obtaining more discriminative tracking features. The detailed structure can be delineated in Fig, \ref{VLMM}.
\par For a given text instance $d_{tk}$ at frame $t$, the coordinates of its four vertices ($x_{i}, y_{i}$) can be obtained from the detector, where $i \in \{1,2,3,4\}$. Simultaneously, the visual feature $F_{lan_{k}} \in R^{c \times d}$ can be acquired from LSC, where $c$ represents the predetermined character length, and $d$ denotes the feature dimension. However, the dimensionality of the position information is $R^{4*2}$, and does not align with the visual feature. Therefore, a transformation is performed to obtain the centerline of text instances. Initially, the lengths of the four edges are computed based on the vertex positions, and the top two edges are selected as the top boundary $tb$ and the bottom boundary $lb$.  Subsequently, $c$ points are sampled from both $tb$ and $lb$, and two sets of points are yielded, namely $P1_{k}=\{tp_{0},tp_{1},...,tp_{i},...,tp_{c-1} \}$ and  $P2_{k}=\{bp_{0},bp_{1},...,bp_{i},...,bp_{c-1} \}$. Finally, the centerline vertexes $P_{c_{k}} \in R^{c \times 2}$ are computed by taking the average of $P1_{k}$ and $P2_{k}$, and normalized using the following equation:
\begin{align}\label{centerline_obtain}
P_{norm_k}=\left\{\left.\left(\frac{x_i}{w_t}, \frac{y_i}{h_t}\right) \right\rvert\, \forall\left(x_i, y_i\right) \in P_{c k}\right\}
\end{align}
where $w_{t}$ and $h_{t}$ represent the width and height of the image in the $t$-th frame, respectively.

\par All normalized centerline $P_{norm_k} \in R^{c \times 2}$ and visual features $F_{lan_{k}} \in R^{c \times d}$ are concatenated along the batch dimension, and are fed into the VPMM. Subsequently, fusion features $F_{fuse}$ are obtained according to Eq. \ref{eq_vlmm}, and are utilized to associate text instances across multiple frames.
 \begin{align}\label{eq_vlmm}
\left\{\begin{array}{l}
\mathrm{F}_{pos}=\varphi_{ {pos }}\left(\varphi_{ {pos }}\left(P_{ {norm }}\right)\right) \\
\mathrm{F}_{ {add }}= { Flatten }\left(\mathrm{F}_{ {pos }}+F_{ {vis }}\right) \\
F_{ {fuse }}=\varphi_{ {fuse }}\left(\varphi_{{fuse }}\left(F_{{add }}\right)\right)
\end{array}\right.
\end{align}
where $\varphi_{pos}$ and $\varphi_{fuse}$ represent one layer multi-layer perceptron (MLP) heads with ReLU activation. $F_{pos} \in R^{n \times c \times d}$ and $F_{lan} \in R^{n \times c \times d}$ represent the position feature after projection and visual feature, respectively. $n$ signifies the number of text instances. Additionally, $F_{add} \in R^{n \times cd}$ and $F_{fuse} \in R^{n \times 2d}$ denote the flattened features and the fusion features, respectively.

\subsection{Training and Inference}
\subsubsection{Model Training}
\par PP-YOLOE-R \cite{ppyolor}, ABINet \cite{ABINet}, and LST-Matcher \cite{GoMatching} are select as the text detector, language synergy classifier (LSC), and text tracker, respectively. Due to the considerable computational overhead and potential optimization conflicts \cite{GoMatching} in the end-to-end training paradigm, all models are trained separately.
\par Initially, PP-YOLOE-R is trained following \cite{ppyolor}. The Varifocal Loss (VFL) $\mathcal{L}_{VFL}$, Probabilistic Intersection-over-Union (ProbIoU) Loss $\mathcal{L}_{ProbIoU}$, and Distribution Focal Loss (DFL) $\mathcal{L}_{DFL}$ are adopted for the optimization of text detector. The detection loss can be defined as follows:
\begin{flalign}
\mathcal{L}_{det}=\alpha \cdot \mathcal{L}_{VFL} + \beta \cdot \mathcal{L}_{ProbIoU} + \gamma \cdot \mathcal{L}_{DFL}
\end{flalign}
where the $\mathcal{L}_{VFL}$, $\mathcal{L}_{ProbIoU}$ and $\mathcal{L}_{DFL}$  are employed for box classification, box regression, and angle prediction, respectively. $\alpha$, $\beta$, $\gamma$ serve as balancing factors.
\par The Varifocal Loss $\mathcal{L}_{VFL}$ can be formulated as follows:
\begin{flalign}
\mathcal{L}_{VFL}=\begin{cases}-q(qlog(p)+(1-q)log(1-p))&q>0\\-\alpha p^\gamma log(1-p)&q=0,\end{cases}
\end{flalign}
where $p$ represents the predicted IoU-Aware Classification Score (IACS) in \cite{VFL}, and $q$ signifies the target score.
\par The $\mathcal{L}_{ProbIoU}$ can be formulated as follows:
\begin{flalign}
\mathcal{L}_{ProbIoU} = \sqrt{1-B_{c}(p,q)}
\end{flalign}
where $B_{c}(p,q)$ denotes the bhattacharyya coefficient in \cite{ProIoU}.
\par The  $\mathcal{L}_{DFL}$ can be formulated as follows:
\begin{flalign}
\mathcal{L}_{DFL}=-\big((y_{i+1}-y)\log(\mathcal{S}_i)+(y-y_i)\log(\mathcal{S}_{i+1})\big).
\end{flalign}
where $y$ represents the target angle.
 $y_i$ and $y_{i+1}$ denote neighboring values around $y$. The variables $\mathcal{S}_i$ and $\mathcal{S}_{i+1}$ refer to the learnable underlying general distributions in \cite{DFL}.
 \par Subsequently, the LSC is trained offline based on the detection outcomes from PP-YOLOE-R. As depicted in Fig. \ref{glyph supervision}, the loss function of the LSC comprises two components: the glyph segmentation loss $\mathcal{L}_{s e g}$, and the original recognition loss $\mathcal{L}_{rec}$ from ABINet. These losses are defined as:
\begin{flalign}
&\
\mathcal{L}_{res}=\lambda_v \mathcal{L}_v+\frac{\lambda_l}{M} \sum_{i=1}^M \mathcal{L}_l^i+\frac{1}{M} \sum_{i=1}^M \mathcal{L}_f^i
\\
&\
    \mathcal{L}_{lsc} = \mathcal{L}_{seg}+\mathcal{L}_{rec}
\end{flalign}
where $\mathcal{L}_{v}$, $\mathcal{L}_{l}$ and $\mathcal{L}_{f}$ represent the cross-entropy losses from visual feature $F_{v}$, language feature  $F_{l}$, and fusion feature $F_{f}$, respectively. The terms $\mathcal{L}_{l}^{i}$ and  $\mathcal{L}_{f}^{i}$ denote the losses at the $i$-th iteration. $\lambda_v$ and $\lambda_l$ stand for balanced factors.
\par Finally, the models from PP-YOLOE-R and ABINet are utilized for the training of LST-Matcher. Given that the spotting model has been fine-tuned on specific datasets, there is no need to optimize
re-scoring head. Thus, only the long short association loss $\mathcal{L}_{asso}$ is adopted to train the text tracker. Specifically, $\mathcal{L}_{asso}$ comprises four components. For each associated trajectory, the losses of ST-Matcher and LT-Matcher represent the log-likelihood of their assignments. The detailed losses are formulated as follows:
 \begin{flalign}
 &\
\mathcal{L}_{s\_{a s s}}\left(E^S, \hat{\tau}_k\right)=-\sum_{t=2}^T \log P_{s_a}\left(\hat{\alpha}_k^t \mid e_{\hat{\alpha}_k^t}^t, E^{S_t}\right)\\
&\
\mathcal{L}_{l\_{a s s}}\left(E^L, \hat{\tau}_k\right)=-\sum_w \sum_{t=1}^T \log P_{l_a}\left(\hat{\alpha}_k^t \mid E_{\hat{\alpha}_k^w}^w, E^L\right)
\end{flalign}
\par For unassociated queries, the ST-Matcher and LT-Matcher will generate empty trajectories. The detailed losses are formulated as follows:
 \begin{flalign}
 &\
\mathcal{L}_{s\_{} b g}\left(E^S\right)=-\sum_{j: \nexists \hat{\alpha}_k^t=j} \sum_{t=2}^T \log P_{s_a}\left(\alpha^t=\emptyset \mid e_j^t, E^{S_t}\right),\\
&\
\mathcal{L}_{l\_b g}\left(E^L\right)=-\sum_{w=1}^T \sum_{j: \nexists \hat{\alpha}_k^w=j} \sum_{t=1}^T \log P_{l_a}\left(\alpha^t=\emptyset \mid E_j^w, E^L\right) .
\end{flalign}
\par Finally, $\mathcal{L}_{asso}$ can be defined as follows:
 \begin{flalign}
\mathcal{L}_{a s s o}=\mathcal{L}_{s\_b g}+\mathcal{L}_{l \_b g}+\sum_{\hat{\tau}_k}\left(\mathcal{L}_{s\_{a s s}}+\mathcal{L}_{l_{\_a s s}}\right) .
\end{flalign}

\subsubsection{Model Inference}
\par As depicted in Fig. \ref{overview}, LOGO conducts video text spotting in an online manner. Initially, LOGO utilizes the rotated detector to localize potential text instances in single frame. Subsequently, text instances with detection score above threshold value are selected, and corresponding text regions in the original images are extracted using RoI Rotate operation. These sampled regions are fed into the Language Synergy Classifier (LSC) for recognition, and language scores are calculated based on recognition results. Following this, fusion scores are computed by taking average of the detection scores and language scores, and are utilized to re-score the detection results prior to tracking. Concurrently, positional information from the detector and visual features from the LSC are fed into the Visual Position Mixture Module (VPMM) to derive fusion features. Ultimately, LST-Matcher associates text instances across multiple frames based on the fusion features of re-scored spotting results.
\section{Experiments}

\begin{figure*}
\centering
\subfigure[GoMatching]{
\begin{minipage}[b]{1\textwidth}
\includegraphics[width=1\textwidth]{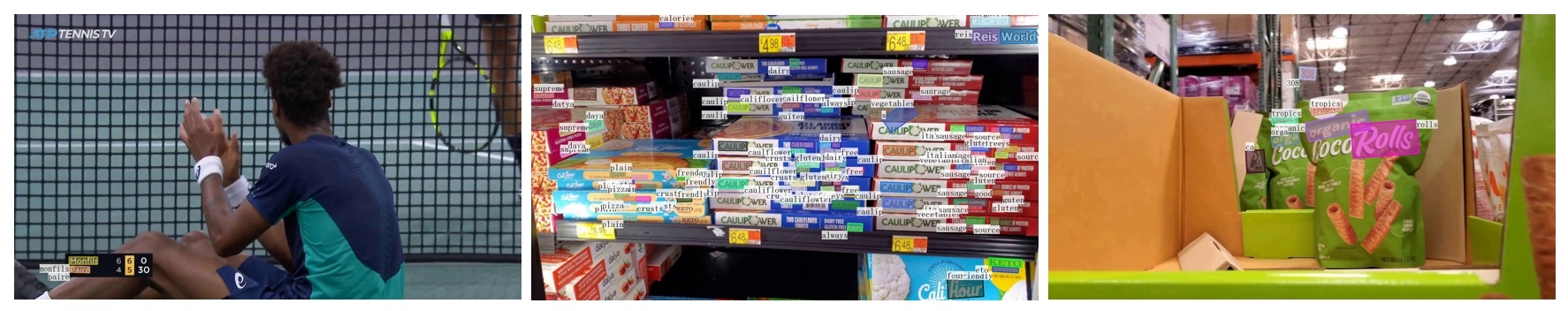}
\end{minipage}
}
\subfigure[LOGO]{
\begin{minipage}[b]{1\textwidth}
\includegraphics[width=1\textwidth]{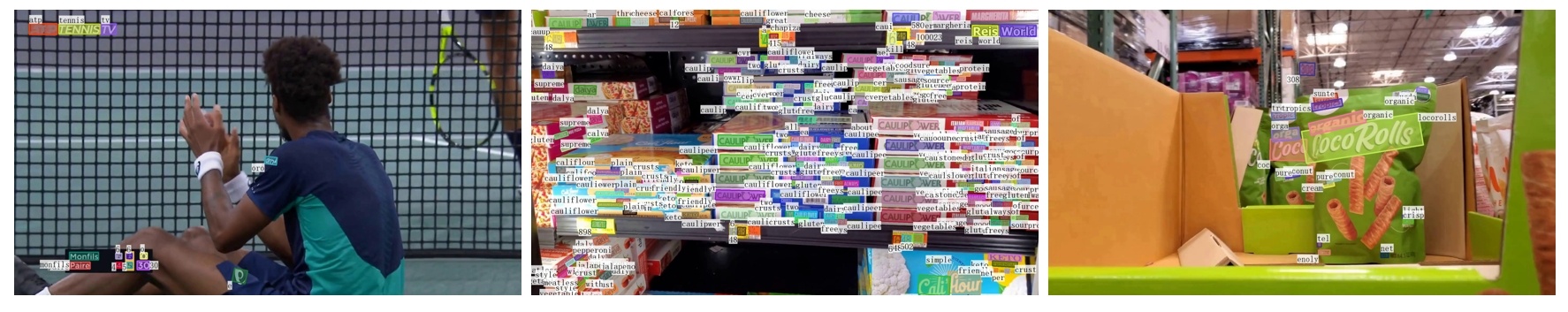}
\end{minipage}
}
\caption{
The visualization of GoMatching and LOGO for video text spotting on DSText \cite{DSText}. While GoMatching tends to overlook noisy text regions to enhance recognition accuracy, LOGO incorporates noisy text regions to retain additional semantic information. } \label{visualize}
\end{figure*}
\subsection{Datasets and Evaluation Metrics}
\textbf{Minetto}. 
The Minetto dataset \cite{mineto} comprises five videos in outdoor scenarios. The resolution of videos is 640$\times$480. All videos are annotated with axis-aligned bounding boxes and serve as testing data. 

 \textbf{ICDAR2013 Video.} The ICDAR2013 Video dataset \cite{ICDAR2013} includes 13 training videos and 15 testing videos in indoor and outdoor scenarios. Text instances in this dataset are labeled as quadrangles with four vertices at the word level.

\textbf{ICDAR2015 Video.} The ICDAR2015 Video dataset \cite{IC5Video} comprises 25 training videos and 24 testing videos.  Text instances are labeled as quadrilateral bounding boxes and transcriptions are provided. 

\textbf{DSText.} The DSText dataset \cite{DSText} comprises 100 videos, 50 videos for training and 50 videos for testing. These videos are harvested from a variety of scenarios (i.e., activities, driving, gaming, sports, indoors, and outdoors). Unlike the ICDAR 2015 Video dataset, the DSText dataset has large proportion of small and dense texts. Therefore, it is more challenging and suitable for evaluating a model's robustness in extreme scenes.

\par \textbf{Evaluation Metrics.} 
In this paper, three tasks are adopted to evaluate the performance of the proposed method. For video text detection, Precision, Recall, and F-measure serve as evaluation metrics. For video text tracking and video text spotting,  Multiple Object Tracking Accuracy (MOTA) \cite{MOTA}, Multiple Object Tracking Precision (MOTP), Identification F1-score (IDF1) \cite{IDF1}, Mostly Tracked (M-Tracked), and Mostly Lost (M-Lost) are employed as evaluation metrics. In this paper, ``Mostly Tracked" (M-Tracked) indicates the number of text instances tracked for at least 80 percent of their lifespan. ``Mostly Lost" (M-Lost) denotes the number of objects tracked for less than 20 percent of their lifespan during tracking. For the Minetto and ICDAR2013 Video, the model trained on the ICDAR2015 Video is utilized for testing purposes.

\begin{table}
\caption{\textnormal{Video text detection performance on ICDAR2013 Video \cite{ICDAR2013}. The optimal and second-best outcomes are highlighted in bold and blue, respectively.}}\label{video_detect}
\centering
\begin{tabular}{c|ccc}
\hline
 \multirow{2}{*}{\makecell[c]{Method}}
 & \multicolumn{3}{c}{Video Text Detection/\%}

 \\
  \cline{2-4}
  &Precision & Recall & F-measure \\ 
 \hline
Epshtein et al, \cite{Epshtein} &39.8&32.5&35.9\\
Zhao et al. \cite{Zhao}&47.0&46.3&46.7\\
Yin et al. \cite{Yin} &48.6&54.7&51.6\\
Khare et al. \cite{Khare}& 57.9&55.9&51.7\\
Wang et al. \cite{Wang}&58.3&51.7&54.5\\
Shivakumara et al. \cite{Shivakumara}& 61.0&57.0&59.0\\
Wu et al. \cite{Wu}& 63.0&68.0&65.0\\
Yu et al. \cite{AGD} &82.4&56.4&66.9\\
Feng et al. \cite{Semantic} &75.5&64.1&69.3\\
Free \cite{Free} &79.7&68.4&73.6\\
CoText \cite{cotext}&82.6&\textbf{71.6}&\textbf{76.7}\\
TransDETR \cite{transdetr}& 80.6&70.2&75.0\\
\rowcolor{mygray}
TransDETR + LSC & \textcolor{blue}{84.3} &\textcolor{blue}{70.2}&\textcolor{blue}{76.6}\\
\rowcolor{mygray}
LOGO & \textbf{85.8}&67.4&75.5\\
\hline
\end{tabular}
\end{table}

\begin{table*}
\centering

\caption{\textnormal{Video text tracking performance on Minetto \cite{mineto}, ICDAR2013 Video \cite{ICDAR2013}, ICDAR2015 Video \cite{IC5Video}, and DSText \cite{DSText}. `M-Tracked' and `M-Lost' represent `Mostly Tracked' and `Mostly Lost', respectively. The integration of the language synergy classifier (LSC) is denoted by `LSC'. `$\dag$' indicate results from the official competition website, while `*' denotes our implementation using official code and weights. `M-ME' signifies the use of multi-model ensemble techniques. The best and second-best results highlighted in bold and blue, respectively.}}   \label{track_result}

\begin{tabular}{c|c|cccccccccccccc}
\hline
 \multirow{2}{*}{Dataset}
 & \multirow{2}{*}{Method}
 & \multicolumn{5}{c}{Video Text Tracking/\%}\\
 \cline{3-7}
 
 & &MOTA$\uparrow$ & MOTP$\uparrow$& $ID_{F1}$$\uparrow$ &   M- Tracked$\uparrow$ & M-Lost$\downarrow$\\ 
 \hline
\multirow{8}{*}{Minetto \cite{mineto}} &
Zuo et al. \cite{Zuo_etal}& 56.4&73.1&-&-&-\\
&Pei at al. \cite{Pei_at_al} &73.1 &57.7&-&-&-\\
&AGD\&AGD \cite{AGD} &75.6&74.7&-&-&-\\
&Yu et al. \cite{AGD} &81.3&75.7&-&-&-\\
&TransVTSpotter \cite{transpotter} &84.1&\textcolor{blue}{77.6}&74.7&-&-\\
&CoText \cite{cotext}& \textcolor{blue}{86.9}&\textbf{80.6}&\textbf{83.9}&87.7&\textbf{0}\\
&TransDETR \cite{transdetr}&86.5&75.5&76.6&\textbf{94.3}&0\\
\rowcolor{mygray} 
\cellcolor{white}&LOGO&\textbf{88.0}&74.1&\textcolor{blue}{80.9}&\textcolor{blue}{92.86}&\textcolor{blue}{0}\\
\hline 

  \multirow{6}{*}{ICDAR2013 Video \cite{ICDAR2013}} &
YORO \cite{YORO} *&47.3&73.7& 62.5&33.1&45.3 \\
&SVRep \cite{SVRep}&53.2&\textbf{76.7}&65.1&38.2&33.2\\
&CoText \cite{cotext}&\textcolor{blue}{55.8}&76.4&\textcolor{blue}{68.1}&44.6&\textbf{28.7}\\
& TransDETR(aug) \cite{transdetr}&54.7&\textcolor{blue}{76.6}&67.2&43.5&33.2\\
\rowcolor{mygray}
\cellcolor{white}
&TransDETR(aug) + LSC &\textbf{56.4}&76.2&\textbf{68.4}&\textbf{48.07}&32.56\\
\rowcolor{mygray}
\cellcolor{white}&LOGO & 55.3 &75.0&67.8&\textcolor{blue}{45.76}&\textcolor{blue}{29.87}\\
 \hline
 \multirow{13}{*}{ICDAR2015 Video \cite{IC5Video}} &
 USTB\_Tex Video \cite{USTB}&7.4&70.8& 25.9&7.4&66.1 \\
 
 & StradVison-1 \cite{USTB} &7.9&70.2&25.9& 6.5&70.8\\
 & USTB(II-2) \cite{USTB}&12.3&71.8&21.9&4.8&72.3\\
 & AJOU \cite{AJOU} &16.4&72.7&36.1&14.1&62.0\\
 &Free \cite{Free}  &43.2&\textcolor{blue}{76.7}&57.9&36.6&44.4\\
 &TransVSpotter \cite{transpotter} &44.1&75.8& 57.3&34.3&33.7\\
 &TextSCM \cite{TextSCM} &44.07&75.19&58.23&44.78&29.02\\
 &Feng at al 
 \cite{Semantic}&48.41&76.31&63.65&36.17&37.73\\
 &CoText \cite{cotext}&51.4&73.6&\textcolor{blue}{68.6}&49.6&\textbf{23.5}\\
  &GoMatching \cite{GoMatching}&\textbf{60.02}&\textbf{77.85}&\textbf{70.85}&\textbf{57.15}&\textcolor{blue}{24.84}\\
 &TransDETR(aug) \cite{transdetr}&47.7 &74.1& 65.5&42.0&32.1\\

\rowcolor{mygray}
\cellcolor{white}& TransDETR(aug) + LSC &48.80&73.52 & 66.99 &
47.96&31.68\\
\rowcolor{mygray}
\cellcolor{white}&LOGO&\textcolor{blue}{55.92}&71.89&68.27&\textcolor{blue}{49.74}&26.51 \\
\hline
\multirow{10}{*}{ DSText \cite{DSText}} &TransVSpotter \cite{transpotter} *&31.49&76.89&44.86&23.17&45.42\\

&TextSCM \cite{TextSCM} *&41.09&76.23&59.00&36.51&44.33\\
 &Feng at al \cite{Semantic} *&39.16&75.56&59.33&38.13&38.47\\
&TransDETR \cite{transdetr} *&38.99&77.15&55.60&37.31&42.57\\
&GoMatching \cite{GoMatching} *&47.41&77.00&60.41&29.05&49.35\\
&TransDETR+HRNet $\dag$&43.52&78.15&62.27&39.60&42.40\\
&DA $\dag$ (M-ME)&50.52&\textcolor{blue}{78.33}&\textcolor{blue}{70.99}&\textcolor{blue}{56.62}&\textcolor{blue}{24.26}\\
&TencentOCR $\dag$ (M-ME)&\textbf{62.56}&\textbf{79.88}&\textbf{75.87}&\textbf{64.51}&\textbf{21.17}\\

\rowcolor{mygray}
\cellcolor{white}&LOGO&\textcolor{blue}{51.36}&77.57&65.70&45.66&37.64 \\
\hline
\end{tabular}
\end{table*}

\subsection{Implementation Details}
\par All experiments are conducted using Tesla V100 GPUs. PP-YOLOE-R \cite{ppyolor}, ABINet \cite{ABINet}, and LST-Matcher \cite{GoMatching} are utilized as the text detector, language synergy classifier (LSC), and text tracker, respectively. PP-YOLOE-R and ABINet are implemented with PaddlePaddle. LST-Matcher employs the official PyTorch code.
\par Specifically, PP-YOLOE-R is pretrained on COCO-TextV2 \cite{cocotext}, and fine-tuned on other video datasets. Images are resized to the maximum target size along the long side. Data augmentation encompasses random flip, random rotation, and batch padding using a stride of 32. For COCO-TextV2, the target size is $1024 \times 1024$ with a batch size of 8. For other datasets, the target size is $1600 \times 900$ with a batch size of 5. Stochastic gradient descent (SGD) with momentum of 0.9 and weight decay of 5e-4 is employed for a total of 36 epochs. A warm-up learning rate strategy with an initial learning rate of 0.008 is adopted for the first 1000 iterations. A cosine learning rate schedule is adopted for subsequent training. The parameters $\alpha$, $\beta$, and $\gamma$ in the detection loss $\mathcal{L}_{det}$ are set to 1.0, 2.5, and 0.05, respectively. Considering that an abundance of high-quality texts are not annotated by ICDAR2015 Video, DeepSolo \cite{deepsolo} is employed to generate pseudo-labels. Pseudo-labels with scores higher than 0.3 and IoU with ground truths lower than 0.2 are added into the original annotations. For DSText, the original annotations are directly used for training.

\par ABINet is initially pretrained on MJSynth (MJ) \cite{MJSynth} and SynthText (ST) \cite{SythText}, and fine-tuned on other video datasets. Images are resized to $128 \times 32$. Data augmentation encompasses geometric transformations (i.e., rotation, affine transformations, and perspective adjustments), image quality degradation and color jitter. The batch size is set to 160, and Adam optimizer with an initial learning rate of 1e-4 is adopted for a total of 10 epochs. The piecewise constant scheduler is employed, and the learning rate decays to 1e-5 after 6 epochs. The parameters $\lambda_v$ and $\lambda_l$
 in the recognition loss $\mathcal{L}_{res}$ are both set to 1, while $h_{I}$ and $h_{S}$ are set to 0.05 and 0.3, respectively. The codes for $\langle EOS \rangle$ and  $\langle PAD \rangle$ are assigned values of 0 and 100, respectively.
\par LST-Matcher is trained based on models from PP-YOLOE-R and ABINet. Following \cite{GoMatching}, a scale-and-crop augmentation strategy is employed. The resolutions in this strategy are set to $1280 \times 1280$ and $1600 \times 1600$ for the ICDAR2015 Video and DSText, respectively. The batch size is set to 6. All frames in a batch originate from the same video. During training, text instances with fusion scores exceeding 0.3 are inputted into the LST-Matcher. During inference, text instances with fusion scores above 0.4 and 0.3 are inputted into LST-Matcher for ICDAR2015 Video and DSText, respectively. AdamW is employed as the optimizer with an initial learning rate of 5e-5. Warmup cosine annealing learning rate scheduler is employed, and the total iterations for ICDAR2015 Video and DSText are set to 30k and 60k, respectively. For DSText, trajectories with scores below 0.6 are excluded in the tracking phase, and noisy trajectories (i.e., recognition results are background or a single character) are removed in the spotting phase. For other datasets, the tracking results from LST-Matcher are directly utilized.

\begin{table*}
\centering
\caption{\textnormal{Video text spotting performance on ICDAR2015 Video \cite{IC5Video}, and DSText \cite{DSText}. `M-Tracked' and `M-Lost' represent `Mostly Tracked' and `Mostly Lost', respectively. The integration of the language synergy classifier (LSC) is denoted by `LSC'. `$\dag$' indicate results from the official competition website, while `*' denotes our implementation using official code and weights. `M-ME' signifies the use of multi-model ensemble techniques. The best and second-best results highlighted in bold and blue, respectively.
}}\label{object metrix}
\label{spotter_result}
\begin{tabular}{c|c|ccccc}
\hline
 \multirow{2}{*}{Dataset}
 & \multirow{2}{*}{Method}
 & \multicolumn{5}{c}{End to End Video Text Spotting/\%}
 \\
 \cline{3-7}
 
 & &MOTA$\uparrow$ & MOTP$\uparrow$& $\mathrm{ID_{F1}}\uparrow$ &   M- Tracked$\uparrow$ & M-Lost$\downarrow$\\ 
 \hline
 \multirow{11}{*}{ICDAR2015 Video \cite{IC5Video}} &
 USTB\_Tex Video(II-2) \cite{USTB}&13.2&66.6& 21.3&6.6&67.7 \\
& USTB\_Tex Video 
 \cite{USTB}&15.6&68.5&28.2&9.5&60.7\\
 &StradViso-1 \cite{USTB}&9.0&70.2&32.0&8.9&59.5\\
 &Free \cite{Free}&53.0&74.9&61.9&45.5&35.9\\
&TransVSpotter \cite{transpotter}&53.2&74.9&61.5&-&-\\
 &TransDETR \cite{transdetr}&58.4&\textcolor{blue}{75.2}&70.4&32.0&20.8\\
 &TransDETR(aug) \cite{transdetr}&60.9&74.6&72.8&33.6&20.8\\
&CoText \cite{cotext}&59.0&74.5&72.0&48.6&26.4\\
&GoMatching \cite{GoMatching} &\textbf{72.04}&\textbf{78.53}&\textbf{80.11}&\textbf{73.30}&\textbf{11.70}\\
\rowcolor{mygray}
\cellcolor{white}
& TransDETR(aug) + LSC&61.62&74.34  &73.96 &
53.91&26.12\\
\rowcolor{mygray}
\cellcolor{white}
&LOGO&\textcolor{blue}{68.07}&73.0&\textcolor{blue}{75.85}&\textcolor{blue}{58.16}&\textcolor{blue}{20.34}\\
\hline
\multirow{8}{*}{DSText \cite{DSText}}
&TextSCM \cite{TextSCM} *&10.74&77.47&42.26&20.24&69.34\\
& TransDETR \cite{transdetr} *&-25.59&\textcolor{blue}{79.88}&24.94&12.05&81.60\\
& Feng at al \cite{Semantic} *&3.09&76.55&37.79&17.95&71.72\\
&GoMatching \cite{GoMatching}& \textcolor{blue}{17.29}&77.48&45.20&19.29&68.65 \\
&abcmot $\dag$ & 5.54 &74.61&24.25&4.20&88.28  \\
&DA $\dag$ (M-ME)& 10.51&78.97&\textcolor{blue}{53.45}&\textcolor{blue}{36.81}&\textcolor{blue}{52.13}\\
&TecentOCR $\dag$ (M-ME)& \textbf{22.44}&\textbf{80.82}&\textbf{56.45}&\textbf{40.25}&\textbf{51.20}\\
\rowcolor{mygray}
\cellcolor{white}
&LOGO&12.94&78.78&46.26&25.75&64.61\\
\hline
\end{tabular}
\end{table*}

\par In light of DSText being a recently published dataset, previous state-of-the-art (SOTA) methods are not trained and tested on it. Therefore, we train and test TransVSpotter \cite{transpotter}, TextSCM \cite{TextSCM}, and Feng et al. \cite{Semantic} based on their official codes. Additionally, due to the absence of recognition modules in TextSCM \cite{TextSCM} and Feng et al. \cite{Semantic},  ABINet \cite{ABINet} is utilized to obtain spotting results. Since TransDETR \cite{transdetr} does not provide results for DSText, the pre-trained model from the official code is utilized to obtain spotting results. Likewise, as tracking results for DSText are unavailable for GoMatching \cite{GoMatching}, a similar approach as with TransDETR is adopted.

\subsection{Comparison with State-of-the-art Methods}

\par To evaluate the effectiveness of LOGO in oriented video texts, comparative analysis  with state-of-the-art methods across three key tasks (i.e., video text detection, video text tracking, and video spotting) is conducted.
\par \textbf{Video text detection.} 
For video text detection, the ICDAR2013 Video is employed to evaluate the detection performance. As is shown in Table \ref{video_detect}, LOGO achieves the highest precision and the third-best F-measure. Additionally, by integrating the language synergy classifier (LSC) into TransDETR, our method achieves a competitive F-measure (76.6\%) compared to the leading approach, CoText \cite{cotext} (76.7\%). This integration leads to improvements of 3.7\% in precision and 1.6\% in F-measure compared to the original model.

\par \textbf{Video text tracking.} 
For video text tracking, Minetto \cite{mineto}, ICDAR2013 Video \cite{ICDAR2013}, ICDAR2015 Video \cite{IC5Video}, and DSText \cite{DSText} are adopted to evaluate the tracking performance. The comparative results are presented in Table \ref{track_result}. Although LOGO only utilizes the CNN-based detector, it achieves the best MOTA and the second-best IDF1 on Minetto. Besides, LOGO achieves a competitive MOTA (55.3\%) compared to the second-best approach, CoText \cite{cotext} (55.8\%) on ICDAR2013 Video. Moreover, by integrating the language synergy classifier (LSC) into TransDETR, our method achieves the best MOTA and IDF1 on ICDAR2013 Video. This integration leads to improvements of 1.7\% and 1.2\% in MOTA and IDF1 compared to the original model. Additionally, without utilizing transformer based text spotters, LOGO achieves notable improvements of 4.52\% in MOTA compared to the best end-to-end method CoText \cite{cotext} on ICDAR2015 Video. In dense and small scenarios, LOGO exhibits improvements of 3.95\%, 5.29\%, and 16.61\% in MOTA, IDF1, and M-Tracked, respectively, compared to the best single-model method GoMatching \cite{GoMatching}. This improvement underscores that relying solely on zero-shot results from DeepSolo may lead to limited tracking trajectories due to domain gap between DSText and normal datasets. While the top-performing method TecentOCR and the third-best method DA leverage multiple model ensembles and large public datasets to enhance tracking performance, LOGO achieves the second-best MOTA (51.36\%) and the third-best IDF1 (65.70\%). This result highlights the significance of language synergy in text tracking.


\begin{table*}
\centering
\caption{\textnormal{ The effect of different components in LOGO for video text tracking task on ICDAR2015 Video \cite{IC5Video} and DSText \cite{DSText}. 'LSC' denotes the language synergy classifier, 'GS' represents glyph supervision, and 'VPMM' refers to the Visual Position Mixture Module.
}}\label{ablation_study_track}
\begin{tabular}{c|c|c|cccccccccccccc}
\hline
\multirow{2}{*}{Dataset}
 & \multirow{2}{*}{Method}
 &\multirow{2}{*}{Pseudo-labels}
 & \multicolumn{5}{c}{Video Text Tracking/\%}\\
   \cline{4-8}

 & & &MOTA$\uparrow$ & MOTP$\uparrow$ & $\mathrm{ID_{F1}}$$\uparrow$  &   M- Tracked$\uparrow$ & M-Lost$\downarrow$\\ 

\hline
\multirow{5}{*}{ICDAR2015 Video \cite{IC5Video}} 
  &Baseline& &41.56&75.15&57.06&24.79&45.46\\
 &Baseline&$\checkmark$  & 49.78&75.49&62.87&37.42&36.17\\
 &Baseline + LSC &$\checkmark$ &52.12&74.90&65.62&42.90&30.48\\
&Baseline + LSC + GS&$\checkmark$  & 52.50&74.90&65.38&42.64&31.16\\
&Baseline + LSC + GS + VPMM &$\checkmark$ &55.92&71.89&68.27&49.74&26.51\\
\hline
\multirow{4}{*}{DSText \cite{DSText}} &
 Baseline&&47.98&79.22& 63.23&38.76&40.38\\
& Baseline + LSC& & 50.33 & 78.82 &64.98&40.78&36.78\\
& Basline + LSC + GS& &50.36&78.83&65.04&40.98&36.57\\
&Baseline + LSC + GS + VPMM& &51.36&77.57&65.70&45.66&37.64\\
\hline
\end{tabular}
\end{table*}


\begin{table*}
\centering
\caption{\textnormal{The effect of different components in LOGO for video text spotting task on ICDAR2015 Video \cite{IC5Video} and DSText \cite{DSText}. 'LSC' denotes the language synergy classifier, 'GS' represents glyph supervision, and 'VPMM' refers to the Visual Position Mixture Module.}}\label{ablation_study_spotter}
\begin{tabular}{c|c|c|ccccccccccccc}
\hline
\multirow{2}{*}{Dataset}
 & \multirow{2}{*}{Method}
  &\multirow{2}{*}{Pseudo-labels}
 & \multicolumn{5}{c}{End to End Text Spotting/\%}

 \\
  \cline{4-8}
 & &&MOTA$\uparrow$ & MOTP$\uparrow$ & $\mathrm{ID_{F1}}$$\uparrow$  &   M- Tracked$\uparrow$ & M-Lost$\downarrow$\\ 
 \hline
   \multirow{5}{*}{ICDAR2015 Video \cite{IC5Video}} 
 &Baseline& &53.19&75.78&64.99&29.55&38.33\\
 &Baseline &$\checkmark$& 61.97&76.14&71.27&46.38&27.07\\
 &Baseline + LSC &$\checkmark$ & 63.39& 75.89&72.85&50.99&22.02\\
 &Baseline + LSC + GS &$\checkmark$ & 64.38&75.89&72.99&50.62&21.58\\
 &Baseline + LSC + GS + VPMM&$\checkmark$ & 68.07&73.00&75.85&58.16&20.34 \\
 \hline
 \multirow{4}{*}{DSText \cite{DSText}} 
& Baseline&&13.57&80.11& 45.13&24.31&63.97\\
& Baseline + LSC& & 14.34& 79.81&45.79&24.08&62.37\\
& Baseline + LSC + GS& & 14.47& 79.84&46.06&24.35&62.53\\
&Baseline + LSC + GS + VPMM& &12.94&78.78&46.26&25.75&64.61\\
\hline
\end{tabular}
\end{table*}

\par \textbf{Video text spotting.}
For video text spotting,  ICDAR2015 Video \cite{IC5Video} and DSText \cite{DSText} are adopted to evaluate spotting performance. The comparative results are presented in Table \ref{spotter_result}. Without relying on  transformer based text spotters, LOGO achieves notable improvements of 7.17\% and 3.05\% in MOTA and IDF1 on ICDAR2015 Video, respectively, compared to the leading end-to-end method TransDETR \cite{transdetr}. For DSText, though multiple model ensembles and large public datasets are utilized by the top-performing method TencentOCR and the fourth-best method DA to enhance spotting performance, LOGO achieves the third-best MOTA (12.94\%) and the third-best IDF1 (46.26\%). However, LOGO's MOTA (12.94\%) falls short of GoMatching's \cite{GoMatching} (17.29\%). To investigate this disparity, further analysis on different tracking thresholds is conducted. As shown in Table \ref{ablation_track_threshold}, LOGO achieves a similar MOTA (17.24\%) when the tracking threshold is set to 0.8. Yet, a high tracking threshold may overlook numerous noisy text regions. Due to external interference, these text regions are partially recognized, but contain valuable semantic information.  Consequently, filtering out these noisy text regions excessively may impede a comprehensive understanding of the video content. Visualization in Fig. \ref{visualize} illustrates that GoMatching tends to disregard noisy text regions to achieve high spotting precision, and corroborates this notion as well.

\subsection{Ablation Studies}
\par Comprehensive ablation studies on ICDAR2015 Video \cite{IC5Video} and DSText \cite{DSText} are conducted to evaluate the effectiveness of the proposed modules. The analysis delves into the impacts of pseudo labels, language synergy classifier (LSC), glyph supervision (GS), and visual position mixture module (VPMM). The evaluation results are presented in Table \ref{ablation_study_track} and Table \ref{ablation_study_spotter}. ByteTrack \cite{bytetrack} is employed as the tracker in baseline, and hand-crafted features is utilized for text association. Additionally, only when the components include VPMM, the LST-Matcher is adopted for text tracking.

 \par \textbf{Effectiveness of Pseudo labels.} Since a substantial number of high-quality text instances are not annotated in ICDAR2015 Video, training on original annotations directly can introduce ambiguity, and degrade model performance. To overcome this dilemma, DeepSolo \cite{deepsolo} is employed for generating pseudo labels. Specifically, pseudo labels with scores exceeding 0.3 and Intersection over Union (IoU) with ground truth lower than 0.2 are added into the original annotations. As shown in Table \ref{ablation_study_track} and Table \ref{ablation_study_spotter}, ICDAR2015 Video is adopted to validate the effectiveness. For video text tracking, the utilization of pseudo labels yields improvements of 8.22\% in MOTA and 5.81\% in IDF1, respectively. Similarly, for video text spotting, leveraging pseudo labels results in improvements of 8.78\% in MOTA and 6.28\% in IDF1, respectively. Thus, although these text instances lack stable trajectories in videos, annotating high-quality texts can reduce detection ambiguity, and significantly fortify model robustness.
\par \textbf{Effectiveness of Language Synergy Classifier.} Since visual features in complex scenarios often pose ambiguity, conventional detection modules struggle to distinguish text instances from background noise. To address this challenge, the language synergy classifier (LSC) is proposed. The LSC integrates both visual context and linguistic knowledge to enhance text spotting performance. The effectiveness of this module is shown in Table \ref{ablation_study_track} and  Table \ref{ablation_study_spotter}.
For video text tracking,  employing the LSC yields improvements of 2.34\% in MOTA and 2.75\% in IDF1 on ICDAR2015 Video, and improvements of 2.35\% in MOTA and 1.75\% in IDF1 on DSText. For video text spotting, leveraging the LSC leads to improvements of 1.42\% in MOTA and 1.58\% in IDF1 on ICDAR2015 Video, and improvements of 0.77\% in MOTA and 0.66\% in IDF1 on DSText. Furthermore, the LSC is integrated into TransDETR to validate its generality. The results after integration are shown in  Table \ref{video_detect}, \ref{track_result}, and  Table \ref{spotter_result}. For ICDAR2013 Video, incorporating the LSC achieves improvements of 3.7\% in precision and 1.6\% in F-measure for video text detection, and improvements of 1.7\% in MOTA and 1.2\% in IDF1 for video text tracking. For ICDAR2015 Video, incorporating the LSC achieves an improvement of  1.1\% in MOTA and 1.49\% in IDF1 for video text tracking, and an improvement of 0.72\% in MOTA and 1.16\% in IDF1 for video text spotting. These gains stem from the powerful ability of LSC in detecting small texts, and filtering out text-like regions.



\begin{table}
\centering
\caption{\textnormal{The effect of Mean Square Error (MSE) loss and Binary Cross Entropy (BCE) loss on ICDAR2015 Video \cite{IC5Video}.}
}\label{ablation_mse_bce}
\begin{tabular}{c|ccccc}
\hline
 \multirow{2}{*}{\makecell[c]{Loss\\ Function}}
 & \multicolumn{5}{c}{End to End Text Spotting/\%}\\
  \cline{2-6}
  &MOTA$\uparrow$ & MOTP$\uparrow$ & $\mathrm{ID_{F1}}$$\uparrow$  &   M- Tracked$\uparrow$ & M-Lost$\downarrow$\\ 
 \hline
BCE &63.46&75.88&72.78&50.91& 21.51\\
MSE&64.38&75.89&72.99&50.62&21.58\\

\hline

\end{tabular}
\end{table}

\begin{table}
\centering
\caption{\textnormal{Video text spotting performance across various tracking thresholds on DSText }
\cite{DSText}.}\label{ablation_track_threshold}
\begin{tabular}{c|ccccc}
\hline
 \multirow{2}{*}{\makecell[c]{Tracking\\ Threshold}}
 & \multicolumn{5}{c}{End to End Text Spotting/\%}

 \\
  \cline{2-6}
  &MOTA$\uparrow$ & MOTP$\uparrow$ & $\mathrm{ID_{F1}}$$\uparrow$  &   M- Tracked$\uparrow$ & M-Lost$\downarrow$\\ 
 \hline
0.6&12.94&78.78&46.26&25.75&64.61\\
0.7&14.09&78.88&46.21&25.48&65.94\\
0.8&17.24&79.33&45.11&23.63&70.88\\
0.82&17.70&79.51&44.20&22,53&72.84\\
0.85&17.59&79.87&41.00&19.43&77.11\\
\hline

\end{tabular}
\end{table}

\par \textbf{Effectiveness of Glyph Supervision.} As shown in Tables \ref{ablation_study_track} and \ref{ablation_study_spotter}, comparative experiments are conducted to demonstrate the effectiveness of glyph supervision. Since recognition accuracy is not the primary concern for video text tracking, and the utilization of glyph supervision may overlook text instances with ambiguous glyph structures, the model exhibits a slight decrease in IDF1 on ICDAR2015 Video, and marginal improvements in MOTA and IDF1 on DSText. Conversely, for video text spotting, employing glyph supervision yields a notable enhancement of 0.99\% in MOTA and 0.14\% in IDF1 on the ICDAR2015 Video, and improvements of 0.13\% in MOTA and 0.27\% in IDF1 on DSText. These evaluation results confirm its effectiveness for noisy text regions. Furthermore, a comparative experiment is conducted on ICDAR2015 Video to analyze the impact of Mean Square Error (MSE) Loss and Binary Cross Entropy Loss (BCE) for video text spotting. As shown in Table \ref{ablation_mse_bce}, the MSE loss yields gains of 0.92\% in MOTA and 0.21\% in IDF1, respectively. Consequently, due to the inherent instability of the pseudo-labels, employing MSE to optimize the network can acquire more robust glyph representations compared to BCE in \cite{glyph_rec}.

\par \textbf{Effectiveness of Visual Position Mixture Module.} The impact of the Visual Position Mixture Module (VPMM) is demonstrated in Table \ref{ablation_study_track} and Table \ref{ablation_study_spotter}. For video text tracking, employing the LST-Matcher with VPMM yields improvements of 3.42\% in MOTA and 2.89\% in IDF1 on the ICDAR2015 Video, and improvements of 1\% in MOTA and 0.66\% in IDF1 on DSText. For video text spotting, the LST-Matcher with VPMM achieves improvements of 3.69\% in MOTA and 2.86\% in IDF1 on ICDAR2015 Video, and a 0.2\% increase in IDF1 on DSText. Regarding the 1.53\% decrease in MOTA on DSText, we attribute this decline to the fact that the LST-Matcher tracks more noisy text instances under the default tracking threshold 0.6. Due to external interference, these noisy text instances pose challenges for the Language Synergy Classifier (LSC) to correctly recognize all characters. To further investigate this, the impact of different thresholds is explored. As illustrated in Table \ref{ablation_track_threshold}, a higher tracking threshold can significantly improve MOTA, and remove noisy text regions. Nonetheless, the partially recognized results still contain valuable semantic information, and overly pursuing high MOTA scores in complex scenarios may impede a deeper understanding of the video content.

\section{Conclusion}
\par In this paper, we propose the Language Collaboration and Glyph Perception Model, termed LOGO. LOGO aims to improve the performance of conventional text spotters by the synergy between visual information and linguistic knowledge. To achieve this goal, a Language Synergy Classifier (LSC) is designed to distinguish text instances from background noise based on the legibility of text regions. The proposed LSC can enhance the scores of text instances with low resolution while diminishing the scores of text-like regions. Additionally, the Glyph Supervision is introduced to further improve recognition accuracy of noisy text regions, and Visual Position Mixture Module is proposed to obtain more discriminative tracking features. Without utilizing transformer based text spotters, LOGO achieves 51.36\% MOTA for video text tracking and 12.94\% MOTA for video text spotting on DSText. Extensive experiments on public benchmarks demonstrate the effectiveness of the proposed method.
\section{Acknowledgement}
\par This work was supported by the Natural Science Foundation of Tianjin (No.21JCYBJC00640) and by the 2023 CCF-Baidu Songguo Foundation (Research on Scene Text Recognition Based on PaddlePaddle).
\bibliographystyle{unsrt}
\bibliography{Mendeley.bib}
\end{document}